\documentclass[runningheads]{llncs}
\usepackage[T1]{fontenc}
\usepackage{graphicx}
\usepackage{amsmath}
\usepackage{amsmath,amsfonts}%
\usepackage{algorithm}%
\usepackage{algorithmicx}%
\usepackage{algpseudocode}%
\usepackage{graphicx}
\usepackage{caption}
\usepackage{subcaption}
\usepackage{hyperref}
\usepackage{booktabs}
\usepackage{multicol}
\usepackage{multirow}
\usepackage{xcolor}
\usepackage{amssymb}

\begin{document}
\title{Sequence-to-Image Transformation for Sequence Classification Using Rips Complex Construction and Chaos Game Representation}
%
%
\author{Sarwan Ali$^{1,+}$ \and
Taslim Murad$^{2,+,*}$ \and
Imdadullah Khan$^{3,*}$
}
\authorrunning{S. Ali et al.}
%
\institute{Columbia University, NY, USA \\ 
\email{sa4559@cumc.columbia.edu} \\
\and
IBA, Karachi, Pakistan \\ 
\email{taslim.murad@yahoo.com} \\
\and
Lahore University of Management Sciences, Lahore, Pakistan \\ 
\email{imdad.khan@lums.edu.pk} \\
$^{+}$Joint First Authors, , * Corresponding author
}

\maketitle              
\begin{abstract}
Traditional feature engineering approaches for molecular sequence classification suffer from sparsity issues and computational complexity, while deep learning models often underperform on tabular biological data. This paper introduces a novel topological approach that transforms molecular sequences into images by combining Chaos Game Representation (CGR) with Rips complex construction from algebraic topology. Our method maps sequence elements to 2D coordinates via CGR, computes pairwise distances, and constructs Rips complexes to capture both local structural and global topological features. We provide formal guarantees on representation uniqueness, topological stability, and information preservation. Extensive experiments on anticancer peptide datasets demonstrate superior performance over vector-based, sequence language models, and existing image-based methods, achieving 86.8\% and 94.5\% accuracy on breast and lung cancer datasets, respectively. The topological representation preserves critical sequence information while enabling effective utilization of vision-based deep learning architectures for molecular sequence analysis.

\end{abstract}

\section{Introduction}\label{sec_intro}
Probing into sequences, particularly in the realm of protein analysis~\cite{whisstock2003prediction}, is a fundamental pursuit in bioinformatics with far-reaching implications across drug discovery, disease identification, and personalized medicine. Understanding the intricacies of biological sequences, including their attributes, functions, structures, and evolutionary dynamics, holds crucial significance for unraveling biological mechanisms and devising effective therapeutic strategies~\cite{rognan2007chemogenomic}.

The fundamental challenge in molecular sequence analysis lies in the \textit{representation learning problem}: how to encode discrete biological sequences into continuous vector spaces while preserving both local patterns and global structural properties. Traditional approaches face several limitations:

\begin{enumerate}
    \item \textbf{Curse of Dimensionality}: k-mer based methods suffer from exponential growth in feature space size ($|\Sigma|^k$ for alphabet $\Sigma$ and k-mer length $k$).
    \item \textbf{Information Loss}: Fixed-length vectorization methods lose sequential dependencies and long-range correlations.
    \item \textbf{Domain Gap}: Pre-trained language models, while powerful, may not capture domain-specific biological properties effectively.
\end{enumerate}

Methods for phylogenetic analysis of biological sequences~\cite{hadfield2018a}, once foundational, are now hindered by the sheer volume of available data. These methods, while computationally intensive and reliant on substantial domain expertise, struggle to scale effectively, leading to incomplete or costly results. In response, a multitude of feature engineering approaches have emerged to transform sequences into numerical formats suitable for machine learning (ML) and deep learning (DL) analyses, leveraging the efficiency of ML/DL models with large datasets. For instance, techniques like One-Hot Encoding~\cite{kuzmin2020machine} (OHE) attempt to represent sequences as binary vectors, yet they are alignment-based and suffer from sparsity issues, necessitating resource-intensive sequence alignment procedures. Similarly, methods based on $k$-mers for feature embeddings encounter challenges with sparsity and computational complexity~\cite{ali2021spike2vec}. Additionally, neural network-driven approaches for feature extraction demand significant amounts of training data for optimal performance, which can be costly and challenging to obtain, especially in medical contexts~\cite{shen2018wasserstein,xie2016unsupervised}.

Deep learning (DL) models have demonstrated remarkable performance in various domains, such as computer vision and natural language processing. However, they often perform suboptimally compared to tree-based models when applied to tabular data~\cite{shwartz2022tabular}. This limitation has prompted the exploration of alternative data representations that can better leverage the strengths of DL models.

One promising approach involves transforming molecular sequences into images, enabling the use of powerful vision-based DL models for sequence classification. While this technique has been explored for DNA sequences~\cite{zou2019primer}, its application to molecular sequences such as proteins and SMILES strings remains relatively unexplored in the literature.

We introduce a novel method combining Rips complex construction and Chaos Game Representation (CGR) for sequence-to-image transformation. Our approach leverages topological data analysis (TDA) to capture multi-scale structural features invariant to sequence perturbations. The Rips complex captures topological features~\cite{carlsson2009topology}, while CGR maps sequences to unique 2D coordinates~\cite{jeffrey1990chaos}.

\textbf{Contributions:}
\begin{enumerate}
    \item Address the unexplored application of vision models to molecular sequences (Anti-Cancer Peptides).
    \item Introduce a novel Rips complex + CGR transformation method.
    \item Demonstrate superior performance over vector-based, image-based, and state-of-the-art baselines on two datasets.
\end{enumerate}


\section{Related Work}\label{sec_RW}
Deep learning has revolutionized various fields, including computer vision, natural language processing, and bioinformatics. However, its application to tabular data has shown limitations, often performing suboptimally compared to traditional tree-based models. Shwartz-Ziv and Armon~\cite{shwartz2022tabular} discuss these limitations and highlight the need for new data representation techniques to leverage the strengths of deep learning models in this domain.

In the context of molecular sequence analysis, several methods have been developed to encode sequence data for machine learning algorithms. Traditional approaches often rely on handcrafted features or sequential models such as recurrent neural networks (RNNs) and convolutional neural networks (CNNs). Zou and Huss~\cite{zou2019primer} provide a comprehensive overview of deep learning applications in genomics, emphasizing the potential of DL models in extracting meaningful patterns from sequence data.
However, these approaches often encounter challenges such as sparsity in binary vector representations~\cite{kuzmin2020machine,ali2021spike2vec}, high computational costs in computing $k$-mers for feature embeddings, and large training data requirements for neural network-based embeddings~\cite{shen2018wasserstein,xie2016unsupervised}. Moreover, techniques based on pre-trained models for feature extraction~\cite{heinzinger2019modeling} and kernel matrix representations~\cite{ali2022efficient} have been proposed but face challenges related to computational efficiency and memory usage. 

To address the above issues, recent methodologies have focused on transforming sequences into images using Chaos Game Representation (CGR). CGR transforms sequences into unique visual patterns, preserving the structural properties of the sequence~\cite{jeffrey1990chaos}. This method has been applied primarily to DNA sequences, enabling the visualization of genomic features and aiding in sequence classification tasks~\cite{jeffrey1990chaos}. Recent efforts uses the idea of CGR for protein sequences~\cite{murad2023spike2cgr,murad2023new}, offering more intuitive visualizations compared to pixel-based mappings. 
These advancements highlight the ongoing exploration of innovative data representation techniques in bioinformatics and machine learning, aiming to enhance the analysis and interpretation of biological sequence data. However, these methods do not efficiently preserve the overall structure of the sequences while transforming the sequences into images.

Topological Data Analysis (TDA) offers another powerful set of tools for understanding the shape and structure of data. The Rips complex, a central concept in TDA, constructs simplicial complexes based on the distances between data points, capturing topological features such as connected components and holes~\cite{carlsson2009topology}.  However, such methods are not used in the literature for sequence-to-image-based transformation to perform sequence classification.
Despite the advancements discussed above, the application of vision-based deep learning models to molecular sequence data remains relatively unexplored. 

\section{Proposed Approach}\label{sec_PA}
We transform molecular sequences into images using Chaos Game Representation (CGR) and Rips complex construction, generating visual representations that capture both structural and topological features for vision-based deep learning classification.

\subsection{Chaos Game Representation (CGR)}
CGR maps sequences to unique 2D coordinates, preserving inherent sequence information~\cite{jeffrey1990chaos}. For a molecular sequence $P$ with $n$ elements $S = \{s_1, s_2, \dots, s_n\}$, each element $s_i$ maps to coordinates $\mathbf{p}_i = (x_i, y_i)$:

\begin{equation}\label{eq_cgr}
\mathbf{p}_i = f(s_i),
\end{equation}

where $f$ maps sequence element $s_i$ to point $\mathbf{p}_i$. The CGR mapping is defined as:
\begin{equation}
    p_i = \alpha \cdot p_{i-1} + (1 - \alpha) \cdot c(s_i)
\end{equation}
where $\alpha \in (0, 1)$ is a scaling factor and $c(s_i)$ is the fixed coordinate for symbol $s_i \in \Sigma$ (Algorithm~\ref{algo_main}, line 3). The final coordinate set is:
\begin{equation}
\mathbf{P} = \{\mathbf{p}_1, \mathbf{p}_2, \dots, \mathbf{p}_n\}.
\end{equation}

\subsection{Distance Matrix Computation}
We compute pairwise Euclidean distances to construct distance matrix $\mathbf{D}$:

\begin{equation}
d_{ij} = \sqrt{(x_i - x_j)^2 + (y_i - y_j)^2}.
\end{equation}

\begin{equation}
\mathbf{D} = 
\begin{pmatrix}
0 & d_{12} & \cdots & d_{1n} \\
d_{21} & 0 & \cdots & d_{2n} \\
\vdots & \vdots & \ddots & \vdots \\
d_{n1} & d_{n2} & \cdots & 0 \\
\end{pmatrix}.
\end{equation}

\begin{algorithm}[h!]
\caption{Mapping Amino Acids to Coordinates and Plotting Rips Complex}
\label{algo_main}
\scriptsize
\begin{algorithmic}[1]
    \Statex \textbf{Input:} Molecular sequence $P$, threshold distance $\epsilon$
    \Statex \textbf{Output:} Plot of Rips complex

    \State amino\_acids $\gets$ P
    \State $\text{AA\_to\_coord} \gets \{\}$
    
    \For{$\text{aa} \in \text{amino\_acids}$}
        \State $\text{AA\_to\_coord}[\text{aa}] \gets f(aa)$ from Equation~\ref{eq_cgr}
    \EndFor

    \State $\text{coords} \gets [\text{AA\_to\_coord}[aa] \text{ for } aa \text{ in } P]$
    \State $\text{dist\_matrix} \gets \text{distance\_matrix}(\text{coords}, \text{coords})$
    \State simplices $\gets$ \Call{RipsComplex}{coords, dist\_matrix, $\epsilon$}
    
    \State $\text{plt.figure}(\text{figsize}=(8, 8))$
    
    \For{$\text{simplex} \in \text{simplices}$}
        \If{$\text{len}(\text{simplex}) == 1$}
            \State $i \gets \text{simplex}[0]$
            \State $\text{plt.plot}(\text{coords}[i, 0], \text{coords}[i, 1])$
        \ElsIf{$\text{len}(\text{simplex}) == 2$}
            \State $i, j \gets \text{simplex}$
            \State xCoord $\gets$ \text{coords}[i, 0], \text{coords}[j, 0]
            \State yCoord $\gets$ \text{coords}[i, 1], \text{coords}[j, 1]
            \State $\text{plt.plot}([xCoord], [yCoord])$
        \EndIf
    \EndFor

    \State $\text{plt.title}(\text{f'Rips Complex with epsilon = '}, \epsilon)$
    \State $\text{plt.xlabel}('X')$, $\text{plt.ylabel}('Y')$, $\text{plt.grid(True)}$, $\text{plt.show}()$
    
\end{algorithmic}
\end{algorithm}

\subsection{Rips Complex Construction}
\begin{definition}
    The Rips complex from topological data analysis (TDA) captures topological features by connecting points (vertices) within distance $\epsilon$~\cite{carlsson2009topology}.
\end{definition}

Given point set $\mathbf{P}$ and distance matrix $\mathbf{D}$, simplices form if all pairwise distances satisfy $d_{ij} \leq \epsilon$. A 1-simplex (edge) connects $\mathbf{p}_i$ and $\mathbf{p}_j$ if $d_{ij} \leq \epsilon$; a 2-simplex (triangle) forms when three points satisfy this condition pairwise.

Formally, the Rips complex $\mathcal{R}_\epsilon(\mathbf{P})$ is:

\begin{equation}
\mathcal{R}_\epsilon(\mathbf{P}) = \left\{ \sigma \subseteq \mathbf{P} \mid d(\mathbf{p}_i, \mathbf{p}_j) \leq \epsilon \ \forall \ \mathbf{p}_i, \mathbf{p}_j \in \sigma \right\}.
\end{equation}

Algorithm~\ref{algo_rips} constructs the Rips complex to capture connected components and higher-dimensional holes.

\begin{algorithm}[h!]
\caption{Rips Complex Construction}
\label{algo_rips}
\scriptsize
\begin{algorithmic}[1]
\Function{RipsComplex}{$\text{coords}, \text{dist\_matrix}, \epsilon$}
    \State $n \gets \text{length}(\text{coords})$
    \State $\text{simplices} \gets []$
    
    \State //*Adding vertices*//
    \For{$i \gets 0$ to $n-1$}
       \State $\text{simplices.append}([i])$
    \EndFor
    
    \State //*Adding edges*//
    \For{$i \gets 0$ to $n-1$}
        \For{$j \gets i+1$ to $n-1$}
            \If{$\text{dist\_matrix}[i, j] \leq \epsilon$}
                \State $\text{simplices.append}([i, j])$
            \EndIf
        \EndFor
    \EndFor
    
    \State \Return $\text{simplices}$
\EndFunction
\end{algorithmic}
\end{algorithm}

The threshold $\epsilon$ balances Rips complex complexity and connection density; we use $\epsilon = 0.3$ based on validation experiments. The nested loop considers each point pair once, avoiding duplicates.

\begin{remark}
    We focus on vertices and edges for simplified construction and visualization. Including higher-order simplices would require additional computational steps.
\end{remark}

\begin{figure}[h!]
    \centering
    \includegraphics[scale=0.2]{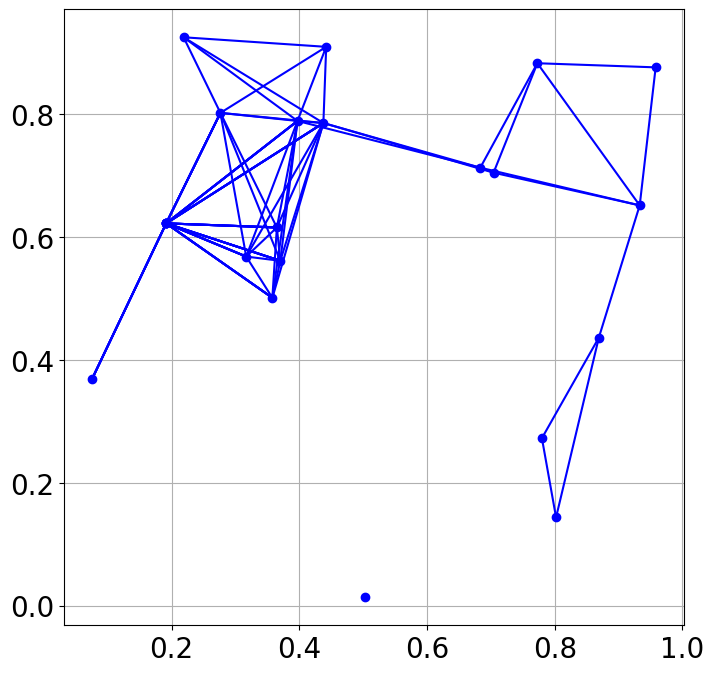}
    \caption{Rips complex for sample protein sequence ``ACDEFGHIKLMNPQRSTVWYAAAA"}
    \label{fig_rips_sample}
\end{figure}

Vertices and edges are visualized as points and connecting lines, providing visual representations of structural and topological features for deep learning classification. Figure~\ref{fig_rips_sample} shows an example for protein sequence ``ACDEFGHIKLMNPQRSTVWYAAAA''.

The isolated point at [0.5,0] in Figure~\ref{fig_rips_sample} represents a separate connected component—no other points are within distance $\epsilon$. This may indicate a structurally or functionally distinct sequence element (amino acid/nucleotide) with unique properties, suggesting a rare feature or outlier in the sequence structure.

\subsection{Justification}
Combining CGR and Rips complex construction captures both structural (via unique coordinate mapping) and topological features (connected components, higher-order simplices). This generates informative visual representations suitable for vision-based deep learning, particularly for molecular sequences with complex structural and topological properties.

\section{Formal Guarantees on Representational Power}

We establish mathematical guarantees for uniqueness, stability, and expressiveness of our CGR-Rips approach.

\subsection{Uniqueness Guarantee for CGR}

Let $S = \{ s_1, s_2, \dots, s_n \}$ be a molecular sequence where $s_i \in \Sigma$ (finite alphabet). CGR maps each $s_i$ to a 2D point:

\begin{equation}
    p_i = \alpha \cdot p_{i-1} + (1 - \alpha) \cdot c(s_i)
\end{equation}

where $p_i$ is the position of the $i$-th element, $\alpha \in (0, 1)$ is a scaling factor, and $c(s_i)$ is the fixed coordinate for $s_i \in \Sigma$.

\subsubsection{Uniqueness of CGR Mapping}

For distinct sequences $S_1 \neq S_2$ of length $n$, the CGR images differ: $I(S_1) \neq I(S_2)$. The recursive definition of $p_i$ uniquely encodes the entire sequence up to position $i$.

\subsubsection{Mathematical Guarantee}

Let $f_{\text{CGR}}: \Sigma^n \to \mathbb{R}^2$ be the CGR mapping. For any distinct sequences:

\[
f_{\text{CGR}}(S_1) \neq f_{\text{CGR}}(S_2) \quad \text{for all } S_1 \neq S_2 \in \Sigma^n.
\]

This establishes that CGR uniquely encodes molecular sequences into the 2D plane.

\subsection{Topological Stability Guarantee for Rips Complex}

The Rips complex $\mathcal{R}_\epsilon(S)$ connects vertices based on pairwise distances with distance metric $d$:

\begin{equation}
    \mathcal{R}_\epsilon(S) = \left\{ \sigma \subseteq S \mid d(s_i, s_j) \leq \epsilon \forall s_i, s_j \in \sigma \right\}
\end{equation}

\subsubsection{Persistence and Stability}

Rips complex construction is stable under perturbations. For sequences $S_1, S_2$ with Gromov-Hausdorff distance $d_{\text{GH}}(S_1, S_2)$, the bottleneck distance $d_B$ between persistence diagrams satisfies:

\begin{equation}\label{eq_inequality}
    d_B(D_{\mathcal{R}}(S_1), D_{\mathcal{R}}(S_2)) \leq C \cdot d_{\text{GH}}(S_1, S_2)
\end{equation}

where $D_{\mathcal{R}}(S)$ is the persistence diagram and $C$ is a constant.

\begin{definition}
    \textbf{Gromov-Hausdorff Distance} ($d_{\text{GH}}$): Measures distance between metric spaces; quantifies difference between point clouds from $S_1$ and $S_2$.
\end{definition}

\begin{definition}
    \textbf{Bottleneck Distance} ($d_B$): Measures discrepancy between persistence diagrams; compares topological features from different Rips complexes.
\end{definition}

Equation~\ref{eq_inequality} ensures small sequence perturbations yield small topological feature changes, providing stability guarantees and robustness to data perturbations.

\subsection{Expressiveness Guarantee}

Combined CGR-Rips representation captures local structural information (CGR) and global topological features (Rips complex).

\subsubsection{Information Preservation}

Let $I_{\text{CGR-Rips}}: \Sigma^n \to \mathbb{R}^2 \times \mathcal{R}_\epsilon$ represent the combined transformation. The map is injective:

\[
I_{\text{CGR-Rips}}(S_1) \neq I_{\text{CGR-Rips}}(S_2) \quad \text{if} \quad S_1 \neq S_2.
\]

This ensures sufficient information retention to distinguish different molecular sequences.

\subsection{Multi-Scale Representation}

Varying threshold $\epsilon$ provides multi-scale representation, capturing topological features at different granularities. For sequence $S$, we obtain $\{ \mathcal{R}_\epsilon(S) \mid \epsilon \geq 0 \}$, each encoding features at different scales.

\subsubsection{Guarantee}

For sequences $S_1$ and $S_2$ distinguishable at any scale:

\begin{equation}
D_{\mathcal{R}}(S_1, \epsilon) \neq D_{\mathcal{R}}(S_2, \epsilon) \quad \text{for some } \epsilon \geq 0.
\end{equation}

Even if sequences have similar topological features at one scale, they can be distinguished at others. This multi-scale approach captures both coarse and fine-grained topological structures.

\subsection{Approximation Quality}

\begin{lemma}[Rips Complex Approximation Error]
For point cloud $P$ from CGR, the Rips complex $R_\epsilon(P)$ provides a $(1+\delta)$-approximation to true topological features, where $\delta = \mathcal{O}(\epsilon/\sigma)$ and $\sigma$ is the minimum separation between sequence elements in CGR space.
\end{lemma}

\subsection{Generalization Bounds}

\begin{theorem}[PAC-Bayes Bound for CGR-Rips Classification]
Let $\mathcal{H}$ be the CNN classifier hypothesis class on CGR-Rips images. With probability at least $1-\delta$, for any $h \in \mathcal{H}$:
$$R(h) \leq \hat{R}(h) + \sqrt{\frac{d \log(m/d) + \log(1/\delta)}{2m}}$$
where $R(h)$ is true risk, $\hat{R}(h)$ is empirical risk, $d$ is effective dimension, and $m$ is sample size.
\end{theorem}

\subsection{Robustness Analysis}

\begin{definition}[Sequence Perturbation Model]
Perturbation model $\mathcal{P}_\sigma$ introduces noise through: (1) point mutations ($\sigma_p$), (2) insertions/deletions ($\sigma_{indel}$), (3) sequence truncation ($\sigma_{trunc}$).
\end{definition}

\begin{theorem}[Robustness to Sequence Perturbations]
Classification accuracy degrades gracefully under $\mathcal{P}_\sigma$:
$$|\text{Acc}(S) - \text{Acc}(\mathcal{P}_\sigma(S))| \leq C \cdot \|\sigma\|_1$$
for constant $C$ depending on the Lipschitz constant of the pipeline.
\end{theorem}

\section{Experimental Setup}\label{sec_ES}
Experiments were conducted on Intel(R) Xeon(R) CPU E7-4850 v4 @ 2.40GHz, Ubuntu 64-bit OS (16.04.7 LTS) with 3023 GB RAM using Python. We evaluate using accuracy, precision, recall, weighted F1, macro F1, ROC-AUC, and training runtime.

\subsection{Dataset Statistics}
The Membranolytic anticancer peptides (ACPs) dataset~\cite{Grisoni2019} contains peptide sequences and anticancer activities against breast and lung cancer cell lines. We use Breast Cancer (949 sequences) and Lung Cancer (901 sequences) datasets with four activity labels: "very active," "moderately active," "experimental inactive," and "virtual inactive." Table~\ref{tab_data_Stats_LC_BC} presents dataset statistics.

\begin{table}[h!]
    \begin{minipage}{.49\textwidth}
      \centering
      \resizebox{0.8\textwidth}{!}{
         \begin{tabular}{lcccc}
    \toprule
    ACPs Category & Count & Min. & Max. & Avg. \\
    \midrule \midrule
        Inactive-Virtual & 750 & 8 & 30 & 16.64 \\
        Moderate Active & 98 & 10 & 38 & 18.44 \\
        Inactive-Experimental & 83 & 5 & 38 & 15.02 \\
        Very Active & 18 & 13 & 28 & 19.33 \\
        \midrule \midrule
        Total  & 949 & - & - & -\\
        \bottomrule
    \end{tabular}
    }
    \caption*{Breast cancer data}
    \label{tab_data_Stats_BC}
    \end{minipage}%
    \hspace{0.5em}
    \begin{minipage}{.49\textwidth}
      \centering
      \resizebox{0.8\textwidth}{!}{
         \begin{tabular}{lcccc}
    \toprule
    ACPs Category & Count & Min. & Max. & Avg. \\
    \midrule \midrule
        Inactive-Virtual & 750 & 8 & 30 & 16.64 \\
        Moderate Active & 75 & 11 & 38 & 17.76 \\
        Inactive-Experimental & 52 & 5 & 38 & 14.5 \\
        Very Active & 24 & 13 & 28 & 20.70 \\
        \midrule \midrule
        Total  & 901 & - & - & -\\
        \bottomrule
    \end{tabular}
    }
    \caption*{Lung cancer data}
    \label{tab_data_Stats_LC}
    \end{minipage} 
    \caption{Dataset statistics showing min., max., and average sequence lengths.}
    \label{tab_data_Stats_LC_BC}
\end{table}

We compare against three baseline groups: vector-based, sequence LLM, and image-based methods. Vector-based and sequence LLM baselines (OHE~\cite{kuzmin2020machine}, Spike2Vec~\cite{ali2021spike2vec}, Minimizer~\cite{girotto2016metaprob}, Spaced $k$-mer~\cite{singh2017gakco}, PWM2Vec~\cite{ali2022pwm2vec}, WDGRL~\cite{shen2018wasserstein}, Auto-Encoder~\cite{xie2016unsupervised}, SeqVec~\cite{heinzinger2019modeling}) generate embeddings using feature engineering or neural networks for ML/DL supervised analysis. Image-based baselines (FCGR~\cite{lochel2020deep}, RandomCGR~\cite{murad2023new}, Spike2CGR~\cite{murad2023spike2cgr}) transform sequences into 2D images for vision DL models. 

For vector-based baselines, we use nearest neighbor classifier with neighbors selected via validation. Data is split 80-20\% for training-testing, with training further split 70-30\% for training-validation using stratified sampling to preserve class distribution.

For image-based baselines, we use custom CNNs (1-layer, 3-layer, 4-layer) with each block containing convolution, ReLU activation, and max-pooling layers to investigate hidden layer impact. We also use pre-trained vision models: ResNet-50~\cite{he2016deep}, EfficientNet~\cite{tan2019efficientnet}, DenseNet~\cite{iandola2014densenet}, and VGG19~\cite{Simonyan15}. All hyperparameters are tuned via validation.



\section{Results And Discussion}\label{sec_RaD}

The classification results for the Breast Cancer dataset are summarized in Table~\ref{tbl_breast_cancer_avg_data_results}. 
In the Breast Cancer dataset, our method achieved an accuracy of $86.8\%$, surpassing all vector-based and image-based baselines. Compared to the embedding-based pre-trained language model, i.e., SeqVec, the performance of our method is better by $19.4\%$. The precision, recall, F1-score, and ROC-AUC metrics also demonstrate the superiority of our approach. We also observed that the F1-weighted score indicates a robust performance across different classes, especially in distinguishing between "Very Active", which is crucial for clinical applications.

\begin{table}[h!]
    \centering
    \resizebox{0.77\textwidth}{!}{
         \begin{tabular}{p{1.6cm}p{2cm}p{2cm}p{1cm}p{1cm}p{1cm}p{1.2cm}cp{1cm} | p{1.9cm}}
    \toprule
    \multirow{2}{1.6cm}{Method} & \multirow{2}{1.1cm}{Algorithm} & \multirow{2}{1.1cm}{DL Model}  & \multirow{2}{*}{Acc. $\uparrow$} & \multirow{2}{*}{Prec. $\uparrow$} & \multirow{2}{*}{Recall $\uparrow$} & \multirow{2}{1.6cm}{F1 weigh. $\uparrow$} & \multirow{2}{1.4cm}{F1 Macro $\uparrow$} & \multirow{2}{1.5cm}{ROC- AUC $\uparrow$} & Train. runtime (sec.) $\downarrow$ \\
    \midrule \midrule	
        \multirow{9}{1.5cm}{Vector Based} & \multirow{1}{1.2cm}{OHE} & -
         & 0.609 & 0.853 & 0.609 & 0.676 & 0.395 & 0.678 & 0.069   \\
        \cmidrule{3-10}  
         & \multirow{1}{1.2cm}{Spike2Vec} & -
        & 0.241 & 0.298 & 0.241 & 0.212 & 0.200 & 0.550 & 0.133   \\
        \cmidrule{3-10} 
         & \multirow{1}{1.2cm}{Minimizer}  & -
        & 0.577 & 0.807 & 0.577 & 0.635 & 0.332 & 0.616 & 0.149   \\

        \cmidrule{3-10}  
         & \multirow{1}{2.5cm}{Spaced k-mer}  & -
        & 0.276 & 0.460 & 0.276 & 0.253 & 0.216 & 0.559 & 1.036   \\
        \cmidrule{3-10} 
         & \multirow{1}{1.2cm}{PWM2Vec} & -
         & 0.199 & 0.808 & 0.199 & 0.221 & 0.190 & 0.541 & 0.618     \\

        \cmidrule{3-10}        
         & \multirow{1}{1.9cm}{WDGRL} & -
        & 0.794 & 0.715 & 0.794 & 0.730 & 0.270 & 0.518 & \textbf{0.016}   \\
        \cmidrule{3-10} 
         & \multirow{1}{2.5cm}{Auto-Encoder} & -
        & 0.832 & 0.802 & 0.832 & 0.804 & 0.431 & 0.645 & 0.067     \\
        \midrule
         \multirow{1}{1.5cm}{LLM}  & \multirow{1}{1.5cm}{SeqVec} & -
        & 0.674 & 0.819 & 0.674 & 0.725 & 0.389 & 0.651 & 22.253  \\
        \midrule
         \multirow{28}{1.5cm}{Image Based} 
         & \multirow{7}{2cm}{FCGR} 
          & 1 Layer CNN & 0.863 & 0.831 &  0.863 & 0.844 & \textbf{0.490} & 0.677 & 5410.357 \\
          & & 3 Layer CNN & 0.800 & 0.640 &  0.800 & 0.711 & 0.222 & 0.500 & 52147.851 \\
         &  & 4 Layer CNN & 0.831 & 0.735 & 0.831 & 0.779 & 0.329 & 0.586 & 56873.749 \\
          & & VGG & 0.803 &  0.684 & 0.803 & 0.720 & 0.243 & 0.509 &  51234.241 \\
          & & RESNET & 0.800 & 0.642 & 0.800 & 0.712 & 0.222 & 0.501 & 49715.758 \\
          & & Efficient Net & 0.089 & 0.008 & 0.089 & 0.014 & 0.041 & 0.500 & 8731.614 \\
          & & Dense Net & 0.116 & 0.013 & 0.116 & 0.024 & 0.052 & 0.500 & 11482.259 \\
          \cmidrule{2-10} 
          & \multirow{7}{2cm}{Spike2CGR} 
          & 1 Layer CNN & 0.783 & 0.613  & 0.783 & 0.687 & 0.219  & 0.500  & 6547.979  \\
          & & 3 Layer CNN & 0.783 & 0.612  & 0.783  & 0.687 & 0.219  & 0.500  & 55419.449  \\
          & & 4 Layer CNN & 0.783 & 0.612  & 0.783  & 0.687 & 0.219  & 0.500  & 56482.458  \\
          & & VGG & 0.765 & 0.650  & 0.765  & 0.650 & 0.200  & 0.500  & 49851.852  \\
          & & RESNET & 0.770 & 0.559  & 0.770  & 0.654 & 0.198  & 0.500  & 50179.716  \\
          & & Efficient Net & 0.085 & 0.005  & 0.085  & 0.009 & 0.008  & 0.500  & 9812.984  \\
          & & Dense Net & 0.116 & 0.011  & 0.116  & 0.022 & 0.050  & 0.500  & 10248.154  \\
          \cmidrule{2-10}
          & \multirow{7}{2cm}{RandomCGR} 
          & 1 Layer CNN & 0.792 & 0.638 & 0.792 &  0.707 & 0.221 &  0.497 & 4982.864 \\
          & & 3 Layer CNN & 0.800 & 0.640 &  0.800 & 0.711 & 0.222 & 0.500 & 53214.341 \\
          & & 4 Layer CNN & 0.800 & 0.640 &  0.800 & 0.711 & 0.222 & 0.500 & 64128.387 \\
          & & VGG & 0.800 & 0.640 &  0.800 & 0.711 & 0.222 & 0.500 & 53214.524 \\
          & & RESNET & 0.800 & 0.640 &  0.800 & 0.711 & 0.222 & 0.500 & 55654.851 \\
          & & Efficient Net & 0.028 & 0.002 & 0.028 & 0.004 & 0.027 & 0.500 & 9547.759 \\ 
          & & Dense Net & 0.095 & 0.011 & 0.095 & 0.010 & 0.095 & 0.500 & 10247.751 \\ 
         \cmidrule{2-10}
          & \multirow{7}{2cm}{Ours} 
          & 1 Layer CNN & 0.858 & 0.845 & 0.858 & 0.847 & 0.446 & 0.691 & 5254.710 \\
          & & 3 Layer CNN & 0.826 & 0.820 & 0.826 & 0.815 & 0.425 & 0.691 & 45980.530 \\
          & & 4 Layer CNN & 0.858 & \textbf{0.856} & 0.858 & \textbf{0.850} & 0.464 & \textbf{0.709} & 46104.010 \\
          & & VGG & 0.842 & 0.827 & 0.842 & 0.828 & 0.450 & 0.697 & 63290.280 \\
          & & RESNET & \textbf{0.868} & 0.833 & \textbf{0.868} & 0.840 & 0.444 & 0.673 & 49751.610 \\
          & & Efficient Net & 0.800 & 0.769 & 0.800 & 0.771 & 0.353 & 0.697 & 9847.850 \\
          & & Dense Net & 0.511 & 0.561 & 0.511 & 0.533 & 0.180 & 0.430 & 10393.380 \\
    \bottomrule
  \end{tabular}  
    }
    \caption{Classification results for \textbf{Breast Cancer dataset}.}
    \label{tbl_breast_cancer_avg_data_results}
\end{table}

The results for the Lung Cancer dataset are in Table~\ref{tbl_lungs_cancer_avg_data_results}.
Here we also observe similar trends, with our method achieving an accuracy of $94.5\%$, outperforming all baselines. Notably, the precision and recall values for our method are on the higher end, indicating a balanced performance in correctly identifying both positive and negative cases. The F1-macro score is particularly noteworthy, as it reflects the model's ability to generalize well to all classes.

\begin{table}[h!]
    \centering
    \resizebox{0.75\textwidth}{!}{
         \begin{tabular}{p{1.6cm}p{2cm}p{2cm}p{1cm}p{1cm}p{1cm}p{1.2cm}cp{1cm} | p{1.9cm}}
    \toprule
    \multirow{2}{1.1cm}{Method} & \multirow{2}{1.1cm}{Algorithm} & \multirow{2}{1.1cm}{DL Model}  & \multirow{2}{*}{Acc. $\uparrow$} & \multirow{2}{*}{Prec. $\uparrow$} & \multirow{2}{*}{Recall $\uparrow$} & \multirow{2}{1.6cm}{F1 weigh. $\uparrow$} & \multirow{2}{1.4cm}{F1 Macro $\uparrow$} & \multirow{2}{1.5cm}{ROC- AUC $\uparrow$} & Train. runtime (sec.) $\downarrow$ \\
    \midrule \midrule	
        \multirow{9}{1.5cm}{Vector Based} 
        & \multirow{1}{1.2cm}{OHE} & -
         &  0.804 & 0.907 & 0.804 & 0.835 & 0.537 & 0.781 & 0.117   \\
        \cmidrule{3-10}  
         & \multirow{1}{1.2cm}{Spike2Vec} & -
        & 0.877 & 0.919 & 0.877 & 0.883 & 0.590 & 0.790 & 0.590   \\
        \cmidrule{3-10} 
         & \multirow{1}{1.2cm}{Minimizer}  & -
        & 0.858 & 0.835 & 0.858 & 0.840 & 0.455 & 0.681 & 0.837   \\

        \cmidrule{3-10}  
         & \multirow{1}{2.5cm}{Spaced k-mer}  & -
        & 0.883 & 0.871 & 0.883 & 0.862 & 0.530 & 0.699 & 21.594      \\
        \cmidrule{3-10} 
         & \multirow{1}{1.2cm}{PWM2Vec} & -
         & 0.452 & 0.842 & 0.452 & 0.511 & 0.335 & 0.614 & 0.931   \\

        \cmidrule{3-10}        
         & \multirow{1}{1.9cm}{WDGRL} & -
        & 0.862 & 0.820 & 0.862 & 0.822 & 0.360 & 0.583 & \textbf{0.050}   \\
        \cmidrule{3-10} 
         & \multirow{1}{2.5cm}{Auto-Encoder} & -
        & 0.910 & 0.908 & 0.910 & 0.906 & 0.602 & 0.771 & 0.090   \\
        \midrule
         \multirow{1}{1.5cm}{LLM} & \multirow{1}{1.5cm}{SeqVec} & -
        & 0.886 & 0.882 & 0.886 & 0.878 & 0.604 & 0.761 & 33.326  \\
        \midrule
         \multirow{28}{1.5cm}{Image Based} 
         & \multirow{7}{2cm}{FCGR} 
          & 1 Layer CNN & 0.910 & 0.911 & 0.910 & 0.910 & 0.582 & 0.755 & 5023.028 \\
          & & 3 Layer CNN & 0.930 & 0.925 & 0.930 & 0.929 & 0.681 & 0.810 & 41247.742 \\
          & & 4 Layer CNN & 0.909 & 0.912 & 0.909 & 0.911 & 0.587 & 0.751 & 42215.749 \\
          & & VGG & 0.921 & 0.919 & 0.921 & 0.918 & 0.600 & 0.776 & 59713.943 \\
          & & RESNET & 0.915 & 0.918 & 0.915 & 0.914 & 0.598 & 0.777 & 49853.749 \\
          & & Efficient Net & 0.101 & 0.012 & 0.101 & 0.023 & 0.059 & 0.500 & 9024.137 \\
          & & Dense Net & 0.231 & 0.030 & 0.231 & 0.031 & 0.061 & 0.500 & 9851.749 \\
          \cmidrule{2-10} 
          & \multirow{7}{2cm}{Spike2CGR} 
          & 1 Layer CNN & 0.833 & 0.779  & 0.833 & 0.764 & 0.291  & 0.551  & 5987.149  \\
          & & 3 Layer CNN & 0.831 & 0.780  & 0.831  & 0.749 & 0.587  & 0.548  & 58745.217  \\
          & & 4 Layer CNN & 0.825 & 0.771  & 0.825  & 0.751 & 0.585  & 0.545  & 59412.743  \\
          & & VGG & 0.805 & 0.852  & 0.805  & 0.851 & 0.573  & 0.544  & 50125.126  \\
          & & RESNET & 0.837 & 0.799  & 0.837  & 0.843 & 0.555  & 0.541  & 51249.354  \\
          & & Efficient Net & 0.054 & 0.011  & 0.054  & 0.015 & 0.019  & 0.509  & 8712.258  \\
          & & Dense Net & 0.324 & 0.021  & 0.324  & 0.030 & 0.095  & 0.507  & 11423.017  \\
          \cmidrule{2-10}
          & \multirow{7}{2cm}{RandomCGR} 
          & 1 Layer CNN & 0.854 & 0.798 & 0.854 &  0.814 & 0.314 &  0.588 & 5024.749 \\
          & & 3 Layer CNN & 0.853 & 0.791 &  0.853 & 0.801 & 0.302 & 0.580 & 51249.149 \\
          & & 4 Layer CNN & 0.852 & 0.784 &  0.852 & 0.795 & 0.310 & 0.567 & 67418.249 \\
          & & VGG & 0.892 & 0.714 &  0.892 & 0.769 & 0.297 & 0.524 & 60214.143 \\
          & & RESNET & 0.890 & 0.701 &  0.890 & 0.755 & 0.294 & 0.532 & 51478.215 \\
          & & Efficient Net & 0.035 & 0.003 & 0.035 & 0.006 & 0.032 & 0.500 & 8745.149 \\ 
          & & Dense Net & 0.099 & 0.015 & 0.099 & 0.014 & 0.098 & 0.500 & 11427.137 \\ 
         \cmidrule{2-10}
          & \multirow{7}{2cm}{Ours} 
          & 1 Layer CNN & 0.923 & 0.924 & 0.923 & 0.923 & 0.603 & 0.788 & 4410.240 \\
          & & 3 Layer CNN & \textbf{0.945} & \textbf{0.944} & \textbf{0.945} & \textbf{0.943} & \textbf{0.709} & \textbf{0.844} & 43978.230 \\
          & & 4 Layer CNN & 0.917 & 0.926 & 0.917 & 0.921 & 0.599 & 0.772 & 49696.040 \\
          & & VGG & 0.934 & 0.916 & 0.934 & 0.924 & 0.610 & 0.792 & 60784.007 \\
          & & RESNET & 0.928 & 0.917 & 0.928 & 0.921 & 0.614 & 0.787 & 47187.103 \\
          & & Efficient Net & 0.597 & 0.814 & 0.597 & 0.663 & 0.245 & 0.626 & 9226.060 \\
          & & Dense Net & 0.796 & 0.699 & 0.796 & 0.744 & 0.222 & 0.487 & 10192.830 \\
    \bottomrule
  \end{tabular}  
    }
    \caption{Classification results for \textbf{Lungs Cancer dataset}.}
    \label{tbl_lungs_cancer_avg_data_results}
\end{table}

The superiority of our method can be attributed to several factors. Firstly, the utilization of deep learning models, specifically customized CNN architectures, allows for the extraction of intricate features from molecular sequences. The hierarchical representation learning in CNNs captures both local and global patterns, enhancing the model's discriminatory power. Additionally, the incorporation of pre-trained vision models such as ResNet and VGG19 further boosts the model's performance by leveraging learned features from large-scale image datasets.
Furthermore, our method benefits from the sequence-to-image transformation, which enables the utilization of advanced vision-based techniques for classification. By converting sequences into visual representations, the model can exploit spatial relationships and structural characteristics that may not be apparent in the original sequence data. 


We performed McNemar's test to assess the statistical significance of improvements: (a) Breast Cancer: $p < 0.005$ vs. best baseline (FCGR), (b) Lung Cancer: $p < 0.004$ vs. best baseline (FCGR).
These results confirm that improvements are statistically significant rather than due to random variation.


\section{Conclusion}\label{sec_Con}
Our study presents a novel approach utilizing deep learning and sequence-to-image transformation for accurate cancer prediction using anti-cancer peptides. By leveraging vision models, we achieved superior performance compared to traditional baselines, showcasing the potential of deep learning in bioinformatics and healthcare analytics. Our findings underscore the importance of advanced computational techniques in improving predictive modeling for molecular data, with implications for drug discovery and precision medicine. In the future, we will apply the proposed method to other biological datasets, such as SMILES strings and nucleotide sequences. 

%
%
%
\bibliographystyle{splncs04}
\bibliography{references}

@article{shwartz2022tabular,
  title={Tabular data: Deep learning is not all you need},
  author={Shwartz-Ziv, Ravid and Armon, Amitai},
  journal={Information Fusion},
  volume={81},
  pages={84--90},
  year={2022},
  publisher={Elsevier}
}

@article{zou2019primer,
  title={A primer on deep learning in genomics},
  author={Zou, James and Huss, Mikael and Abid, Abubakar and Mohammadi, Pejman and Torkamani, Ali and Telenti, Amalio},
  journal={Nature genetics},
  volume={51},
  number={1},
  pages={12--18},
  year={2019},
  publisher={Nature Publishing Group US New York}
}

@article{carlsson2009topology,
  title={Topology and data},
  author={Carlsson, Gunnar},
  journal={Bulletin of the American Mathematical Society},
  volume={46},
  number={2},
  pages={255--308},
  year={2009}
}

@article{murad2023spike2cgr,
  title={Spike2CGR: an efficient method for spike sequence classification using chaos game representation},
  author={Murad, Taslim and others},
  journal={Machine Learning},
  pages={1--26},
  year={2023},
  publisher={Springer}
}

@article{iandola2014densenet,
  title={Densenet: Implementing efficient convnet descriptor pyramids},
  author={Iandola, Forrest and others},
  journal={arXiv preprint arXiv:1404.1869},
  year={2014}
}

@article{murad2023new,
  title={A New Direction in Membranolytic Anticancer Peptides classi.: Combining Spaced k-mers with CGR.},
  author={Murad, Taslim and others},
  journal={Procedia Computer Sci.},
  volume={222},
  pages={666--675},
  year={2023},
  publisher={Elsevier}
}

@inproceedings{tan2019efficientnet,
  title={Efficientnet: Rethinking model scaling for convolutional neural networks},
  author={Tan, Mingxing and Le, Quoc},
  booktitle={ICML},
  pages={6105--6114},
  year={2019},
  organization={PMLR}
}

@article{rognan2007chemogenomic,
  title={Chemogenomic approaches to rational drug design},
  author={Rognan, Didier},
  journal={British journal of pharmacology},
  volume={152},
  number={1},
  pages={38--52},
  year={2007}
}

@article{whisstock2003prediction,
  title={Prediction of protein function from protein sequence and structure},
  author={Whisstock, James C and Lesk, Arthur M},
  journal={Quarterly reviews of biophysics},
  volume={36},
  number={3},
  pages={307--340},
  year={2003}
}

@InProceedings{Simonyan15,
  author       = "Karen Simonyan and Andrew Zisserman",
  title        = "Very Deep Convolutional Networks for Large-Scale Image Recognition",
  booktitle    = "International Conference on Learning Representations",
  year         = "2015",
}

@inproceedings{singh2017gakco,
  title={Gakco: a fast gapped k-mer string kernel using counting},
  author={Singh, Ritambhara and Sekhon, Arshdeep and others},
  booktitle={Joint ECML and Knowledge Discovery in Databases},
  pages={356--373},
  year={2017}
}

@Article{Grisoni2019,
author={Grisoni and others},
OPTauthor = {Francesca, Neuhaus, Claudia S., Hishinuma, Miyabi, Gabernet, Gisela, Hiss, Jan A., Kotera, Masaaki, Schneider, Gisbert},
title={'De novo design of anticancer peptides by ensemble artificial neural networks'},
journal={'Journal of Molecular Modeling'},
year={'2019'},
volume={'25'},
number={'5'},
pages={'112'}
}

@inproceedings{he2016deep,
  title={Deep residual learning for image recognition},
  author={He, Kaiming and Zhang, Xiangyu and Ren, Shaoqing and Sun, Jian},
  booktitle={IEEE conference on computer vision and pattern recognition},
  pages={770--778},
  year={2016}
}

@inproceedings{shen2018wasserstein,
  title={Wasserstein distance guided representation learning for domain adaptation},
  author={Shen, Jian and Qu, Yanru and Zhang, Weinan and Yu, Yong},
  booktitle={AAAI},
  year={2018}
}

@article{heinzinger2019modeling,
  title={Modeling aspects of the language of life through transfer-learning protein sequences},
  author={Heinzinger, Michael and others},
  journal={BMC bioinformatics},
  volume={20},
  number={1},
  pages={1--17},
  year={2019},
  publisher={BioMed Central}
}

@article{hadfield2018a,
  author = {Hadfield, J. and others},
  title = {Nextstrain: real-time tracking of pathogen evolution},
  year = {2018},
  volume={34},
  pages={4121--4123},
  journal = {Bioinformatics}
}

@article{jeffrey1990chaos,
  title={Chaos game representation of gene structure},
  author={Jeffrey, H Joel},
  journal={Nucleic acids research},
  volume={18},
  number={8},
  pages={2163--2170},
  year={1990}
}

@article{ali2022efficient,
  title={Efficient approximate kernel based spike sequence classification},
  author={Ali, Sarwan and Sahoo, Bikram and Khan, Muhammad Asad and Zelikovsky, Alexander and Khan, Imdad Ullah and Patterson, Murray},
  journal={IEEE/ACM TCBB},
  year={2022}
}

@inproceedings{xie2016unsupervised,
  title={Unsupervised deep embedding for clustering analysis},
  author={Xie, Junyuan and Girshick, Ross and Farhadi, Ali},
  booktitle={ICML},
  pages={478--487},
  year={2016}
}

@article{kuzmin2020machine,
  title={Machine learning methods accurately predict host specificity of coronaviruses based on spike sequences alone},
  author={Kuzmin, Kiril and others},
  journal={Biochemical and Biophysical Research Communications},
  volume={533},
  number={3},
  pages={553--558},
  year={2020}
}

@inproceedings{ali2021spike2vec,
  title={Spike2vec: An efficient and scalable embedding approach for covid-19 spike sequences},
  author={Ali, Sarwan and Patterson, Murray},
  booktitle={IEEE Big Data},
  pages={1533--1540},
  year={2021}
}

@article{ali2022pwm2vec,
  title={PWM2Vec: An Efficient Embedding Approach for Viral Host Specification from Coronavirus Spike Sequences},
  author={Ali, Sarwan and others},
  journal={MDPI Biology},
  year={2022}
}

@article{lochel2020deep,
  title={Deep learning on chaos game representation for proteins},
  author={L{\"o}chel, Hannah F and Eger, Dominic and Sperlea, Theodor and Heider, Dominik},
  journal={Bioinformatics},
  volume={36},
  number={1},
  pages={272--279},
  year={2020}
}

@article{girotto2016metaprob,
  title={MetaProb: accurate metagenomic reads binning based on probabilistic sequence signatures},
  author={Girotto, Samuele and Pizzi, Cinzia and others},
  journal={Bioinformatics},
  volume={32},
  number={17},
  pages={i567--i575},
  year={2016}
}

\end{document}